\documentclass[letterpaper]{article} 
\usepackage{aaai2026}  
\usepackage{times}  
\usepackage{helvet}  

\usepackage{courier}  
\usepackage[hyphens]{url}  
\usepackage{graphicx} 
\usepackage{natbib}  
\usepackage{caption} 
\usepackage{amsmath,amssymb}
\usepackage{booktabs}
\usepackage{xcolor}
\usepackage{adjustbox}
\usepackage{multirow}
\usepackage{makecell}
\usepackage{makecell,array,booktabs,xcolor}

\usepackage{algorithm}
\usepackage{algorithmic}

\usepackage{newfloat}
\usepackage{listings}
\DeclareCaptionStyle{ruled}{labelfont=normalfont,labelsep=colon,strut=off} 
\floatstyle{ruled}
\newfloat{listing}{tb}{lst}{}
\floatname{listing}{Listing}
\title{When CNNs Outperform Transformers and Mambas: Revisiting Deep Architectures for Dental Caries Segmentation}
\author{
    Aashish Ghimire\textsuperscript{\rm 1\equalcontrib}
, Jun Zeng\textsuperscript{\rm 3\equalcontrib}, Roshan Paudel\textsuperscript{\rm 1}, Nikhil Kumar Tomar\textsuperscript{\rm 1}, Deepak Ranjan Nayak\textsuperscript{\rm 5}, Harshith Reddy Nalla\textsuperscript{\rm 1}, Vivek Jha\textsuperscript{\rm 4}, Glenda Reynolds\textsuperscript{\rm 2}, Debesh Jha\textsuperscript{\rm1}\thanks{Corresponding author: debesh.jha@usd.edu.} 
}
\affiliations{
    \textsuperscript{\rm 1}Department of Computer Science, University of South Dakota, USA\\
    \textsuperscript{\rm 2}Department of Dental Hygiene, University of South Dakota, USA\\
    \textsuperscript{\rm 3}School of Software Engineering, Chongqing University of Posts and Telecommunications, China\\
    \textsuperscript{\rm 4}Post Graduate Institute of Medical Education \& Research, Chandigarh, India\\
  \textsuperscript{\rm 5}Malaviya National Institute of Technology, Jaipur, India

}

\usepackage{bibentry}

\begin{document}

\maketitle

\begin{abstract}
Accurate identification and segmentation of dental caries in panoramic radiographs are critical for early diagnosis and effective treatment planning. Automated segmentation remains challenging due to low lesion contrast, morphological variability, and limited annotated data. In this study, we present the first comprehensive benchmarking of convolutional neural networks, vision transformers and state-space mamba architectures for automated dental caries segmentation on panoramic radiographs through a DC1000 dataset. Twelve state-of-the-art architectures, including VMUnet, MambaUNet, VMUNetv2, RMAMamba-S, TransNetR, PVTFormer, DoubleU-Net, and ResUNet++, were trained under identical configurations. Results reveal that, contrary to the growing trend toward complex attention based architectures, the CNN-based DoubleU-Net achieved the highest dice coefficient of 0.7345, mIoU of 0.5978, and precision of 0.8145, outperforming all transformer and Mamba variants. In the study, the top 3 results across all performance metrics were achieved by CNN-based architectures.  Here, Mamba and transformer-based methods, despite their theoretical advantage in global context modeling, underperformed due to limited data and weaker spatial priors. These findings underscore the importance of architecture-task alignment in domain-specific medical image segmentation more than model complexity. Our code is available at: \url{https://github.com/JunZengz/dental-caries-segmentation}.


\end{abstract}


\section{Introduction}

Dental caries is among the most common chronic diseases worldwide~\cite{abdalla2022deep, lee2021automated}. Although largely preventable, dental caries remains a leading cause of tooth loss and oral discomfort~\cite{hirata2023deep}. Epidemiological data show that untreated dental caries affects over one-third of the global population. In children, tooth decay is the most common chronic dental condition, impacting approximately 514 million individuals worldwide~\cite{gbd2018study2017}. Therefore, early and accurate detection of carious lesions is crucial. Conventional diagnostic methods, such as visual-tactile examination, intraoral radiography, approximal tooth separation, caries-detection dyes, fiber-optic transillumination (FOTI), and indices such as DMFT, are widely used. These approaches, however, are limited by subjective interpretation, inter-examiner variability, low sensitivity to early enamel lesions, false-positive staining, and difficulties in detecting early-stage or overlapping lesions~\cite{abdalla2022deep,srilatha2019diagnosticaids,abdelaziz2023earlycaries}.

\begin{figure}[t!]
\centering
\includegraphics[width=0.45\textwidth]{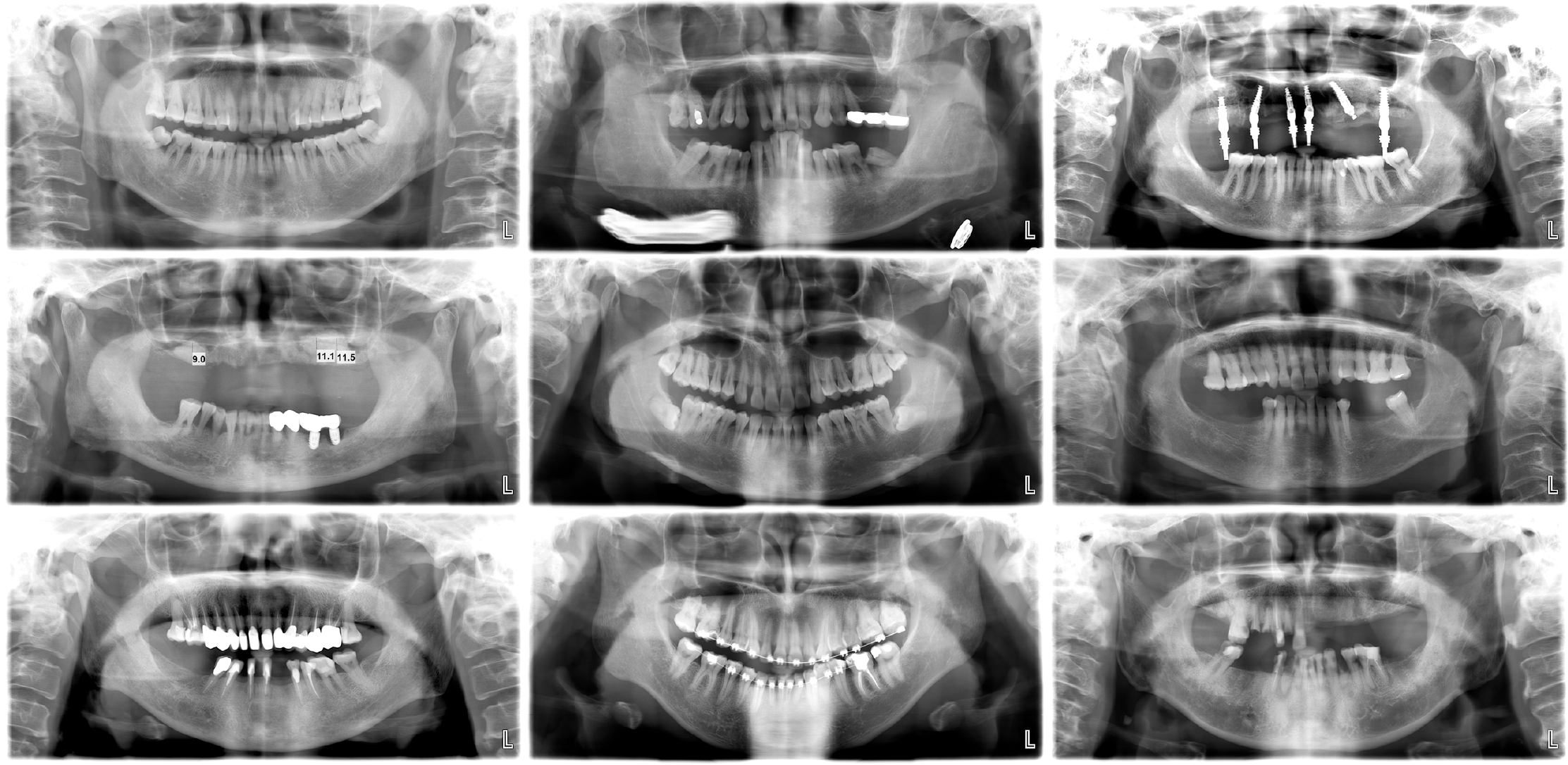}
\caption{Variation in dental panoramic X-ray structures across the DC1000 dataset, showing patients with full dentition, missing teeth, and braces.}
\label{fig:dc1000_variations}
\end{figure}

\begin{table*}[!t]
\centering
\scriptsize
\setlength{\tabcolsep}{8pt} 
\renewcommand{\arraystretch}{2}

\caption{Representative studies on DL approaches for dental segmentation. Reported performance corresponds to each study’s best model configuration.}

\resizebox{\textwidth}{!}{
\begin{tabular}{p{3cm} p{2.5cm} p{3.3cm} p{3.5cm} p{4.6cm}}
\toprule
\textbf{Paper (Year)} & \textbf{Modality} & \textbf{ Architecture} & \textbf{Dataset (Size)} & \textbf{Results} \\
\midrule

Park~et~al. (2022), \textit{BMC Oral Health}~\cite{park2022caries} &
\makecell[l]{Teeth /\\intraoral photographs} &
\makecell[l]{U-Net + ResNet-18 +\\Faster R-CNN(Combined)} &
\makecell[l]{In-house dataset: 2,348 images\\from 445 patients} &
\makecell[l]{\textbf{Segmentation-based DL} improved\\accuracy 0.813, AUC 0.837,\\Precision 0.874, Sensitivity 0.890.} \\

\midrule

Zhu~et~al. (2022), \textit{Neural Comput. Appl.}~\cite{zhu2023cariesnet} &
\makecell[l]{Teeth /\\panoramic radiographs} &
\makecell[l]{CariesNet (U-shape CNN\\with FSAA module)} &
\makecell[l]{1,159 images, 3,217\\annotated caries lesions} &
\makecell[l]{\textbf{CariesNet} reached Dice 93.64\% and\\accuracy 93.61\% for caries detection.} \\

\midrule

Asci~et~al. (2024)~\cite{asci2024deeplearning} &
\makecell[l]{Teeth / Pediatric\\panoramic radiographs} &
\makecell[l]{U-Net (CNN)} &
\makecell[l]{6,075 images (ages 4–14,\\mixed dentition)} &
\makecell[l]{\textbf{U-Net} achieved Sensitivity 0.827,\\Precision 0.912, F1-score 0.867.} \\

\midrule

Hao~et~al. (2024)~\cite{hao2024semisupervised} &
\makecell[l]{Teeth /\\Panoramic radiographs} &
\makecell[l]{SemiTNet (Transformer-\\based, semi-supervised)} &
\makecell[l]{TSI15k: 1,589 labeled +\\14,728 unlabeled ($\sim$16k total)} &
\makecell[l]{\textbf{SemiTNet} achieved IoU 94.41\% and\\Dice 95.45\% leveraging unlabeled data.} \\

\midrule

Lim~et~al. (2025), \textit{Med. Image Anal.}~\cite{lim2025dualcnn} &
\makecell[l]{Teeth /\\Panoramic radiographs} &
\makecell[l]{Dual-CNN: Faster R-CNN +\\U-Net (EfficientNet-B0)} &
\makecell[l]{597 images (443 train, 50 val,\\100 test)} &
\makecell[l]{\textbf{Dual-CNN} obtained Dice 0.743, \\ IoU 0.608, Recall 0.731, Precision 0.788.} \\

\midrule

Damrongsri~et~al. (2025), \textit{Clin. Oral Imaging}~\cite{pornprasertsuk2025clinical} &
\makecell[l]{Teeth /\\Panoramic radiographs} &
\makecell[l]{YOLOv5 + Attention U-Net} &
\makecell[l]{500 images, 14,997 teeth\\(clinically validated GT)} &
\makecell[l]{\textbf{Attention U-Net} - Dice 0.85, IoU 0.75,\\ Recall 0.96, Precision 0.77, Accuracy 0.93.} \\

\bottomrule
\end{tabular}}
\vspace{2pt}
\footnotesize
\\
\textit{Abbreviations:} FSAA – Full-Scale Axial Attention; GT – Ground Truth
\label{tab:caries_benchmark_studies}
\end{table*}

Deep neural networks have transformed medical image analysis in recent years. Convolutional Neural Networks (CNNs) have been especially successful in segmentation tasks (e.g., U-Net~\cite{ronneberger2015unet} and its variants), due to their ability to learn rich hierarchical features from images. More recently, Vision Transformers (ViTs)~\cite{dosovitskiy2020vit,chen2021transunet} with self-attention mechanisms have emerged, showing promise in capturing global context dependencies and improving long-range reasoning. In parallel, a new family of state-space models has emerged and its medical adaptations, such as VM-UNet~\cite{ruan2024vmunet}, VM-UNetV2~\cite{zhang2024vm} and Rmamba-s~\cite{zeng2025reverse}, aim to balance efficiency and global modeling through selective state-space recurrence.

There are studies in the literature that have exhibited strong performance. But most of them only use CNNs and do not systematically compare with Transformers, and Mamba-based models. In this field, it remains unclear whether architectural complexity truly translates into better performance for dental applications.  To study the research question, we perform a comprehensive benchmark on the DC1000 dataset to rigorously study different segmentation architectures under the same experimental settings. Figure~\ref{fig:dc1000_variations} shows the structural variability across the dataset, showcasing cases with full dentition, missing teeth, and orthodontic appliances. These examples highlight the real-world challenges of panoramic radiographs, where anatomical variation, metallic artifacts, and inconsistent contrast complicate automatic caries detection. The main contributions of this work are as follows:

\begin{itemize}
    \item \textbf{Comprehensive benchmark:} This work presents a comprehensive benchmark evaluation of 12 segmentation architectures, spanning from CNNs, transformers, and Mamba-based models, on the DC1000 panoramic radiograph dataset for automated caries segmentation on the DC1000 dataset.

    \item \textbf{Results:} Our analysis reveals contrasting results, where convolutional networks consistently outperform Transformers and Mamba-based methods in dice coefficient, mIoU, precision and inference speed. Here, DoubleU-Net achieved the highest dice coefficient of 0.7345 and mIoU of 0.5978, while Mamba-based RMAMamba-S achieved mDSC of 0.6583 and transformer-based PVTFormer achieved the mDSC of 0.6733, establishing a comprehensive benchmark for dental caries segmentation. The  9\% higher mDSC with significantly fewer parameters and faster inference highlights the importance of spatial inductive priors in data-limited medical imaging.

    \item \textbf{Clinical relevance and translational potential:} With this work, we highlight the relevance and translational potential of this problem into computer-aided dental diagnosis. We identify some of the strengths and limitations of current algorithms with qualitative and quantitative results, informing clinicians about key considerations in adopting AI-based diagnostic systems.
    
\end{itemize}

\section{Related Works}

Deep Learning is one of the fundamental components that aid automated diagnosis of dental imaging, especially for segmenting dental caries in panoramic X-rays.  Initial research on the field has focused on convolutional architectures such as U-Net~\cite{ronneberger2015unet} and its variants. For instance, Park et al. (2022)~\cite{park2022caries} used a U-Net to segment the tooth surface from Intraoral photograph and then passed the segmented output to the ResNet-18~\cite{he2015deepresiduallearningimage} for caries image classification, and utilized Faster R-CNN~\cite{ren2015faster} for the localization of carious lesions with improved accuracy from 0.758 to 0.813, and AUC from 0.731 to 0.837. Similarly, Asci et al.(2024)~\cite{asci2024deeplearning} used a U-Net model on more than 6000 pediatric radiographs, indicating consistent performance across primary, mixed, and permanent dentitions. Moreover, to tackle the challenge of automated diagnosis in panoramic x-rays, Hamamci et al. (2023)~\cite{hamamci2023dentexabnormaltoothdetection} introduced the DENTEX benchmark, which is the publicly available hierarchically annotated dataset to support abnormal tooth detection through panoramic X-rays. However, it does not focus on segmentation tasks, and evaluation relies on metrics like AP/AR, which have known limitations in medical contexts such as class imbalance sensitivity, and interpretability.

Recently, attention mechanisms and transformer-based architectures were introduced in the field to enhance the identification of small or low-contrast carious lesions. Zhu et al.~\cite{zhu2023cariesnet} proposed CariesNet, a U-shaped network with full-scale axial attention that outperformed conventional CNNs in multi-stage caries segmentation. Hao et al.~\cite{hao2024semisupervised} developed a semi-supervised transformer model called SemiTNet, which achieved state-of-the-art IoU and dice scores by utilizing unlabeled panoramic radiographs. These experiments show a growing transition from traditional CNN-based segmentation approaches to attention-driven and hybrid methods. In 2025, Lim et al.~\cite{lim2025dualcnn} integrated Faster R-CNN~\cite{ren2015faster} with U-Net to improve lesion-level recall. On the other hand, Pornprasertsuk-Damrongsri et al.(2025)~\cite{pornprasertsuk2025clinical} used a two-stage YOLOv5 + Attention U-Net pipeline, validated against bitewing-confirmed data, to achieve high recall for posterior lesions.

Table~\ref{tab:caries_benchmark_studies} summarizes representative DL studies on dental image segmentation on various image modalities, detailing their datasets, model types, and reported performance. Furthermore, many studies employ differing protocols and do not offer standardized toolkits or unified benchmarks, thereby limiting reproducibility and generalization. Additionally, the literature indicates that benchmarking efforts on the publicly available DC1000 dataset still remain sparse and lack systematic cross-architectural evaluation. To address these gaps, we present a first unified benchmark of 12 diverse DL models, including CNNs, transformer-based, and Mamba-based architectures. All of our models are evaluated under a consistent training pipeline. Our study aims to serve as a foundational resource guiding the development of robust, generalizable and clinically relevant caries segmentation models.

 \section{METHOD}
Most of the medical image segmentation methods still depends on CNNs as their underlying foundation. The number of convolutional layers with a gradual reduction of the spatial resolution helps to extract local and global information. Architectures that use encoder-decoder structure like U-Net~\cite{ronneberger2015unet}, ResUNet++~\cite{jha2019resunetplusplus}, DoubleU-Net~\cite{jha2020doubleu} and ColonSegNet~\cite{jha2021real} are enhanced with skip connection, and thus are able to reconstruct minute detail. In addition to that, they can learn to increase abstract semantic representation. These designs can be explained as being characterized by effective training dynamics, stability on heterogeneous data, and high spatial resolution, which makes them useful in detecting small or low-contrast caries.

Transformer-based models replace localized convolutional processing with self-attention and in this way, allow the network to learn long-range interaction across an image. Canonical models include TransNetR, TransRUPNet, RSAFormer, and PVTFormer divide the image in smaller patches that are then described as tokens, and hierarchical feature constructions is made possible with multiple attention layers \cite{jha2024transnetr, jha2024transrupnet, yin2024rsaformer, jha2024ct}. These models are good in the context of the global context, but they usually demand large amounts of training data and large amounts of computational resources. As a result, they may not be able to reach their full potential in the application to small groups of medical imaging when the fine-scale information within the signal prevails.

The recent mamba-based architectures, such as VMUNet~\cite{ruan2024vm},  VMUNetV2~\cite{zhang2024vm}, MambaUNet~\cite{wang2024mamba}, and RMAMamba-S~\cite{zeng2025reverse}, provide a different paradigm of visual processing, which makes use of state-space modeling. They do not use attention but instead convert information into a sequence of sequential state transitions that enables the efficient management of long-range dependencies at a lower computing cost. The visual state Mamba model is tested in the current study as a lightweight option to transformer-based architectures. Although it shows some scaling potential, panoramic dental image performance is limited by the small size of the dataset used and the subtle changes in texture depending on the carious area, which highlights the importance of additional domain-specific adaptation.

\begin{figure}[t]
\centering
\includegraphics[width=\columnwidth]{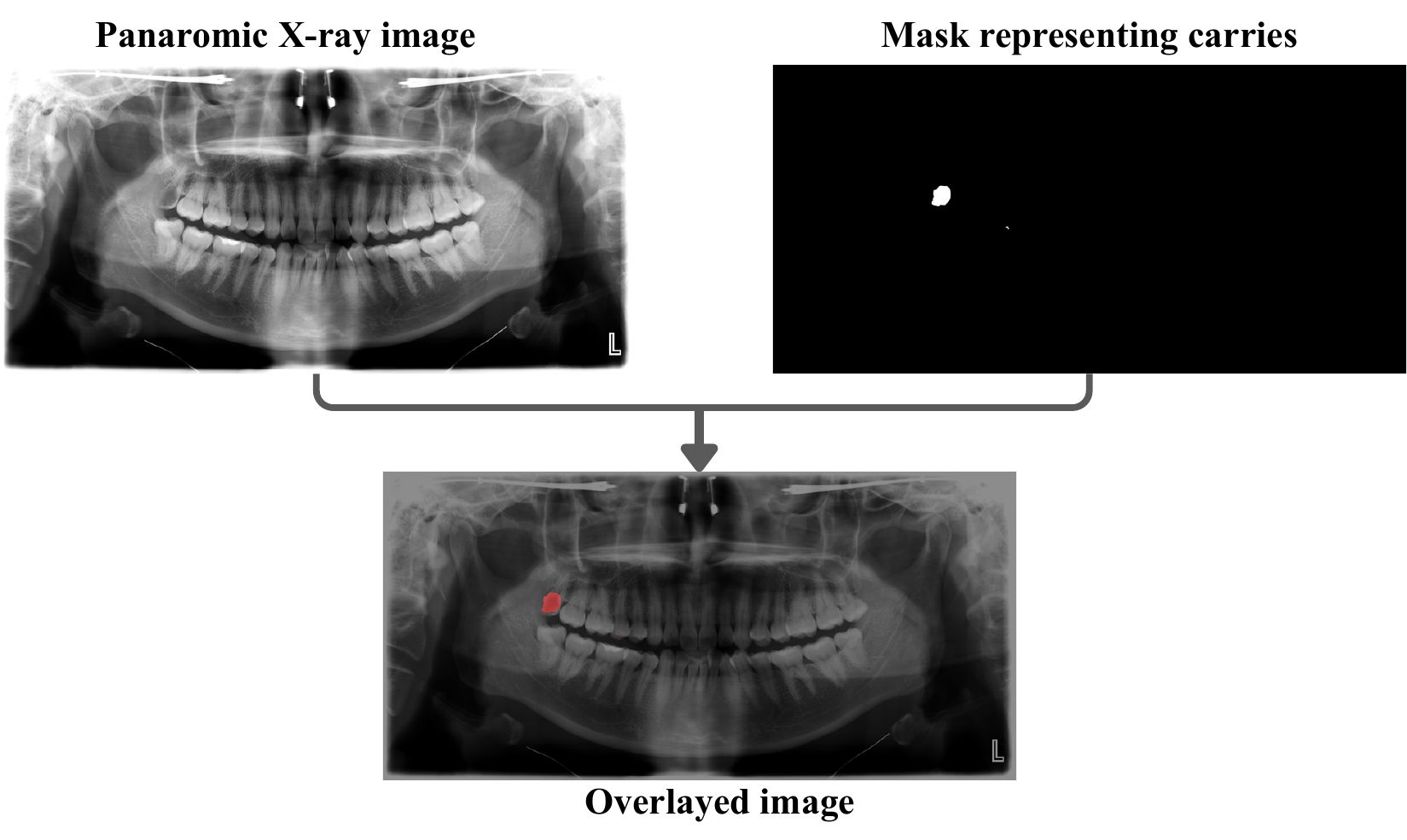}
\caption{Illustration of dental caries annotation in the DC1000 dataset. The figure shows the original panoramic X-ray image, the expert-annotated binary mask highlighting carious regions, and the overlay visualization combining both. 
}
\label{fig:overlayed_description}
\end{figure}


\section{Experimental setup }

\begin{table*}[t!]
\centering
\caption{Performance comparison of 12 segmentation models on the DC1000 dataset. 
Top three results are highlighted in \textcolor{red}{\textbf{red}} (best), 
\textcolor{blue}{\textbf{blue}} (second), and \textcolor{green}{\textbf{green}} (third).}
\resizebox{\textwidth}{!}{
\begin{tabular}{c|l|ccccccc} 
\hline
\textbf{Type} & 
\textbf{Model} & \textbf{mIoU ($\uparrow$)} & \textbf{mDSC ($\uparrow$)} & \textbf{Recall ($\uparrow$)} & \textbf{Precision ($\uparrow$)} & \textbf{F2 ($\uparrow$)} & \textbf{HD ($\downarrow$)} \\
\hline

\multirow{4}{*}{Mamba-based} & 
VMUNet~\cite{ruan2024vm} & 0.4652 & 0.6114 & 0.5981 & 0.6899  & 0.6003 & 2.2505 \\

 & VMUNetV2~\cite{zhang2024vm} & 0.4039 & 0.5549 & 0.5516 & 0.6308  & 0.5475 & 2.3620 \\

 & MambaUNet~\cite{wang2024mamba} & 0.4302 & 0.5759 & 0.6019 & 0.6503  & 0.5861 & 2.3653 \\

 & RMAMamba-S\cite{zeng2025reverse} & 0.5124 & 0.6583 & 0.6366 & 0.7348  & 0.6424 & 2.2225  \\
\hline

\multirow{4}{*}{Transformer-based} & 
RSAFormer~\cite{yin2024rsaformer} & 0.4465 & 0.5916 & 0.5880 & 0.6825  & 0.5863 & 2.3834  \\

& TransNetR~\cite{jha2024transnetr} & 0.4946 & 0.6400 & 0.6052 & 0.7411  & 0.6157 & 2.2513  \\

& TransRUPNet~\cite{jha2024transrupnet} & 0.5278 & 0.6723 & 0.6460 & 0.7631  & 0.6526 & 2.1715  \\

& PVTFormer~\cite{jha2024ct} & 0.5291 & 0.6733 & 0.6564 & 0.7461 & 0.6594 & 2.1637 \\
\hline

\multirow{4}{*}{CNN-based} &
ResUNet++\cite{jha2019resunetplusplus} & 0.5197 & 0.6637 & 0.6463 & 0.7308  & 0.6503 & 2.1420   \\

& ColonSegNet~\cite{jha2021real} & \textcolor{green}{\textbf{0.5669}} & \textcolor{green}{\textbf{0.7044}} & \textcolor{green}{\textbf{0.6819}} & \textcolor{blue}{\textbf{0.7779}}  & \textcolor{green}{\textbf{0.6877}} & \textcolor{green}{\textbf{2.0686}}   \\

& U-Net~\cite{ronneberger2015unet} & \textcolor{blue}{\textbf{0.5836}} & \textcolor{blue}{\textbf{0.7236}} & \textcolor{red}{\textbf{0.7159}} & \textcolor{green}{\textbf{0.7649}} & \textcolor{red}{\textbf{0.7161}} & \textcolor{red}{\textbf{2.0022}}  \\

& DoubleU-Net~\cite{jha2020doubleu} & \textcolor{red}{\textbf{0.5978}} & \textcolor{red}{\textbf{0.7345}} & \textcolor{blue}{\textbf{0.7009}} & \textcolor{red}{\textbf{0.8145}} & \textcolor{blue}{\textbf{0.7115}} & \textcolor{blue}{\textbf{2.0260}}   \\

\hline
\end{tabular}}
\label{tab:dc1000_results}
\end{table*}

\subsection{Dataset}

In this study, we have used DC1000 dataset, which consists of 597 high resolution panoramic images. Each of them were annotated with pixel-level segmentation masks for dental caries\cite{wang2023multi}. These radiographs were annotated by experienced dentists, and were obtained from the clinical sources, such that it ensures dataset's reliability and quality~\cite{lee2021automated}. In addition to that, this dataset was collected from a wide range of population, as they have variation in caries size, intensity, and location, making it suitable for segmentation task~\cite{ma2021benchmarking}. In this dataset, we have 497 training images, out of which 4 radiographs do not have mask, and 100 test images. However, this dataset do not provide any official validation split.  Images in this dataset are stored in 8-bit grayscale format, and its corresponding masks are encoded as a binary map. Notations like 0 and 1 are used in this data, where 0 represents background and 1 denotes carious lesions. To maintain the balance of data, we have used the provided training set, and optionally, split it further into training and validation subsets at 80:20 ratio.

\subsection{Experimental setup and configuration}

To ensure consistency in runtime, throughput, and reproducibility, all models were trained on the PyTorch framework using a single NVIDIA V100 GPU with 32GB memory. Furthermore, the performance of all models was assessed using standard segmentation metrics, including mean Intersection over Union (mIoU), Dice coefficient(mDSC),  Precision, Recall and F2-score. To ensure a fair comparison across architectures, all models were trained and evaluated under a unified experimental configuration using identical data splits, augmentations, and optimization settings. All experiments were conducted using a consistent training pipeline to ensure fair comparison across models.

\subsection{Data augmentation and preprocessing}
Data augmentations included horizontal flipping, random shifts, rotations, mild scaling, contrast-limited adaptive histogram equalization, and brightness–contrast adjustment. Validation images were only resized and normalized. Model performance was evaluated after each epoch.  The best model checkpoint was selected based on the highest validation IoU to prevent overfitting. All metrics and losses were logged for quantitative analysis. Results were plotted to visualize convergence trends across different architectures. The training configuration is summarized as follows:

\begin{itemize}
    \item \textbf{Optimizer:} Adam 
    \item \textbf{Loss:} Binary Cross-Entropy with logits (pos\_weight = 18.0) + Focal loss ($\alpha=0.75$, $\gamma=2.0$) combined in a 0.5 : 0.5 ratio
    \item \textbf{Learning Rate:} $1\times10^{-4}$ with ReduceLROnPlateau scheduler (${\text{patience}} = $ 5)
    \item \textbf{Epochs:} 500 with early stopping based on validation Dice (patience = 50)
    \item \textbf{Batch Size:} 4
    \item \textbf{Input Size:} $384\times384$ (3-channel RGB panoramics)
    \item \textbf{Data Augmentations:}
    \begin{itemize}
        \item Resize to $384\times384$
        \item Random horizontal flip ($p=0.5$)
        \item Shift, scale, and rotation (shift $\pm5\%$, scale $\pm5\%$, rotate $\pm15^{\circ}$, border = constant, $p=0.5$)
        \item Random brightness and contrast adjustment ($p=0.3$)
    \end{itemize}

\end{itemize}

\section{Results and discussion}

\begin{figure*}[t!]
\centering
\makebox[\textwidth][c]{%
    \includegraphics[width = \textwidth]{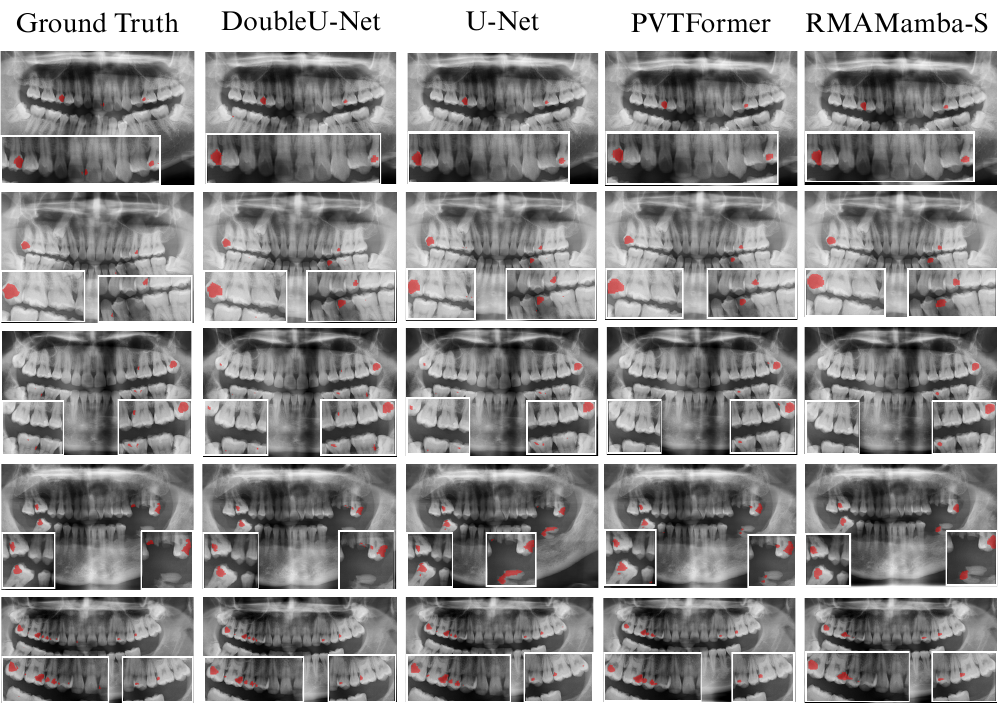}
}
\caption{ Qualitative comparison of segmentation outputs from four representative architectures on the DC1000 test set. The first column shows the ground truth masks, followed by predictions from DoubleU-Net, U-Net, PVTFormer, and RMAMamba-S. DoubleU-Net consistently yields the most accurate lesion outlines, preserving fine radiolucent boundaries and diffuse enamel–dentin transitions that other models often miss.}
\label{fig:qualitative_result}
\end{figure*}

\begin{table*}[t!]
\centering
\scriptsize
\setlength{\tabcolsep}{7pt} 
\caption{Computational complexity and efficiency comparison of the evaluated segmentation models on the DC1000 dataset. Lower GFLOPs and fewer parameters indicate higher computational efficiency. The top three FPS values are highlighted in \textcolor{red}{\textbf{red}}, \textcolor{blue}{\textbf{blue}}, and \textcolor{green}{\textbf{green}}.}
\begin{tabular}{p{4.45cm} p{3.85cm} >{\centering\arraybackslash}p{0.6cm} >{\centering\arraybackslash}p{0.4cm} >{\centering\arraybackslash}p{0.4cm} p{5.35cm}}
\toprule
\textbf{Model} & \textbf{Architecture Type} &\textbf{GFLOPs} ($\downarrow$) & \textbf{Params (M)} ($\downarrow$) & \textbf{FPS} ($\uparrow$) & \textbf{Efficiency Insight} \\
\midrule
VMUnet~\cite{ruan2024vm}        & \makecell[l]{Classical encoder--decoder Mamba}                 & \textcolor{red}{\textbf{9.25} }             & 22.04                     & 25.17 & Low GFLOPs; moderate accuracy. \\
VMUnetV2~\cite{zhang2024vm}     & \makecell[l]{Vision Mamba UNet with SDI block}                 & \textcolor{blue}{\textbf{9.90}}     & 17.91                     & 21.02 & Complex vision mamba design; poor accuracy. \\
MambaUnet~\cite{wang2024mamba}  & \makecell[l]{UNet-like pure visual Mamba}                      & \textcolor{green}{\textbf{10.35}} & \textcolor{green}{\textbf{15.48}} & 26.24 & Baseline Mamba; low parameters. \\
RMAMamba-S~\cite{zeng2025reverse} & \makecell[l]{Reverse-attention Vision Mamba}                  & 27.95                      & 55.21                     & 11.43 & High parameters; moderate performance. \\
RSAFormer~\cite{yin2024rsaformer} & \makecell[l]{Attention-enhanced dual decoder}                 & 31.69                      & 65.76                     &  9.95 & High complexity but poor performance. \\
TransNetR~\cite{jha2024transnetr} & \makecell[l]{Encoder--decoder hybrid Transformer}             & 22.70                      & 15.03                     & 31.17 & High GFLOPs; transformer-heavy. \\
TransRUPNet~\cite{jha2024transrupnet} & \makecell[l]{Transformer--residual U-Net hybrid}            & 89.40                      & 25.64                     & 23.82 & High computational cost; moderate accuracy. \\
PVTFormer~\cite{jha2024ct}      & \makecell[l]{Hierarchical transformer encoder}                 & 97.64                      & 45.51                     & 16.96 & Large model; powerful global reasoning. \\
ResUNet++~\cite{jha2019resunetplusplus} & \makecell[l]{Residual U-Net with SE blocks}                & 35.58                      & \textcolor{red}{\textbf{4.06}}             & \textcolor{blue}{\textbf{34.22}} & Lightweight; strong efficiency with moderate accuracy. \\
ColonSegNet~\cite{jha2021real}  & \makecell[l]{Lightweight CNN encoder--decoder}                 & 139.86                     & \textcolor{blue}{\textbf{5.01}}    &  2.31 & Computationally heavy due to multi-branch decoding. \\
U-Net~\cite{ronneberger2015unet} & \makecell[l]{Classical encoder--decoder CNN}                 & 147.43                     & 34.53                     & \textcolor{red}{\textbf{40.83}} & Baseline CNN; higher FLOPs despite simple structure. \\
DoubleU-Net~\cite{jha2020doubleu} & \makecell[l]{Stacked dual- decoder CNN}                      & 121.40                     & 29.29                     & \textcolor{green}{\textbf{33.09}} & Balanced trade-off between accuracy and cost. \\
\bottomrule
\end{tabular}
\label{Table3}
\end{table*}

\begin{figure}[!t]
	\centering
	\includegraphics[width=\linewidth]{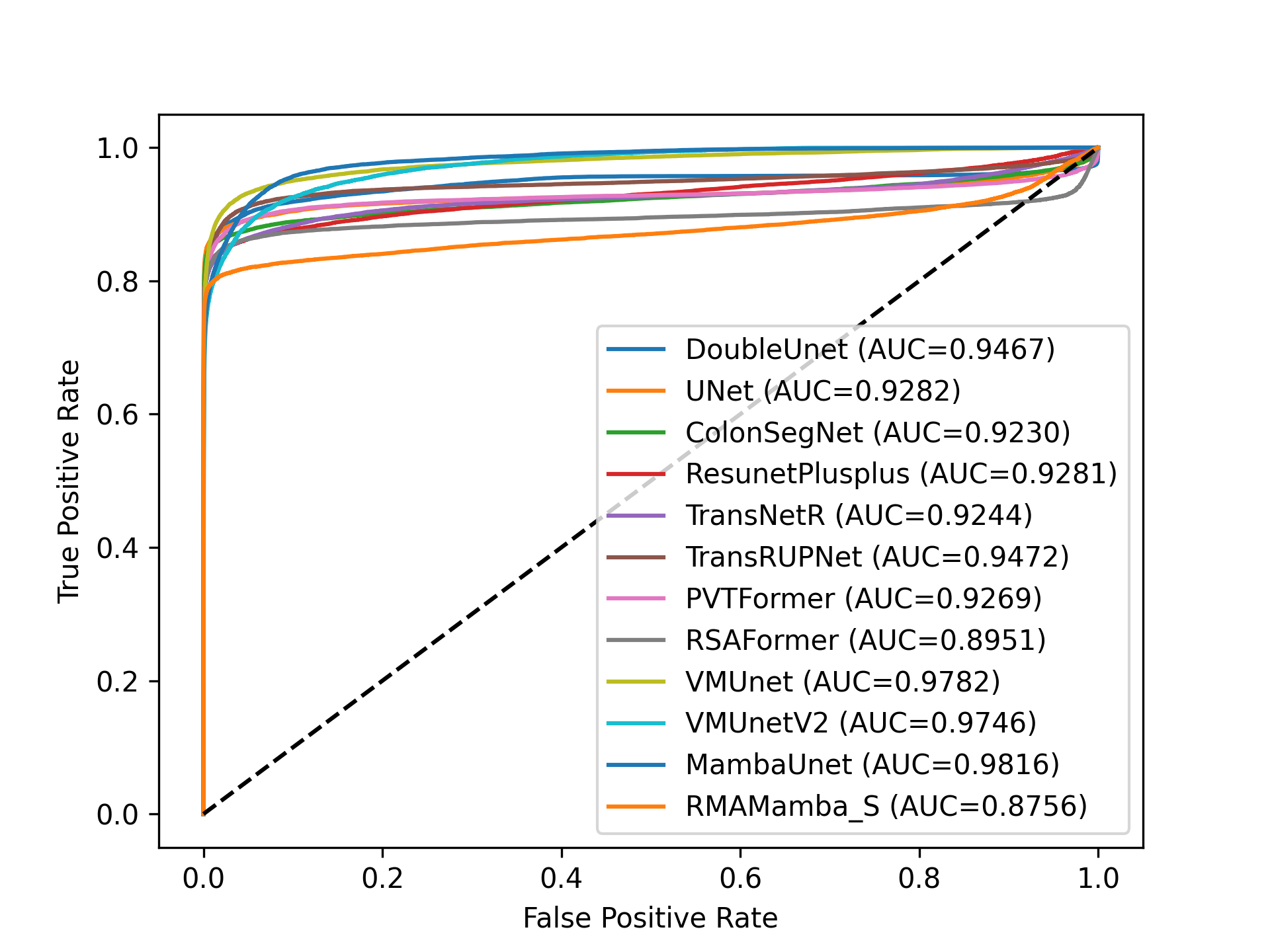}
	\caption{ROC curves comparing pixel-level discriminative performance of several CNN, Transformer, and Mamba architectures.}
        \label{fig:roccurvecomparision}
\end{figure}

\subsection{Quantitative results}

As seen in Table~\ref{tab:dc1000_results}, DoubleU-Net achieved the highest segmentation performance over other models, with an mIoU of 0.5978, mDSC of 0.7345, Recall of 0.7009, and Precision of 0.8145 This indicates strong consistency between pixel level overlap and boundary description, highlighting the advantage of its dual-decoder design in capturing minute carious regions. Despite being a 2020 convolutional neural network architecture, DoubleU-Net outperformed more recent transformer-based approaches, suggesting that efficient multi-scale feature aggregation can be highly competitive for dental caries segmentation. Additionally, U-Net ranked second overall  with mIoU of 0.5836, mDSC of 0.7236, having balanced recall of 0.7159 and precision of 0.7649, while maintaining a low boundary error, HD of 2.0022. This balance makes U-Net favorable for real-time or near–real-time diagnostic settings. In contrast, ColonSegNet achieved the third-best performance in terms of mIoU of 0.5669  and recall of 0.6819, also showing a third-lowest boundary deviation with HD of 2.0945, suggesting a potential trade-off between segmentation quality and contour precision.

The carious region covers just a fraction of each panoramic radiographs, and hence it amplifies the impact of missing pixels on the mIoU and Dice coefficients. To some extent, this sparsity explains the recall versus precision differences between the models and it highlights the sensitivity of the boundary based measures, such as the Hausdorff Distance.
Between the transformer-based models, PVTFormer has the best scores with mIoU of 0.5291, mDSC of 0.6733, closely followed by TransRUPNet with a mIoU of 0.5278 and mDSC of 0.6723, while RSAFormer delivered a slightly lower mIoU of 0.4465 with the highest HD of 2.3834. While discussing about the state-space models, the RMAMamba achieved the best in terms of mIoU of 0.5124 and better precision and recall with the lowest boundary error with HD of 2.2225 among Mamba architectures. To conclude, CNN-based models currently provide better segmentation results, transformer-based models maintain a balance with enhanced global context reasoning, and Mamba models has a compromise between accuracy and efficiency, showing the strengths and trade-offs of each approach.

\subsection{Qualitative results}
Figure~\ref{fig:qualitative_result} shows qualitative results comparison four representative algorithms DoubleUNet~\cite{jha2020doubleu}, UNet~\cite{ronneberger2015unet}, PVTFormer~\cite{jha2024ct} and RMAMamba-S~\cite{zeng2025reverse} due to limited space. DoubleU-Net consistently provides the most accurate delineation of carious lesions, preserving both subtle radiolucent boundaries and diffuse enamel–dentin transitions. U-Net shows strong structural localization but slightly under-segments low-contrast regions. PVTFormer captures broader lesion context yet occasionally over-segments due to sensitivity to intensity variations and metallic artifacts. RMAMamba-S produces smoother, less precise boundaries and frequently misses fine lesion details. From a clinical perspective, accurate boundary localization is critical for assessing lesion severity and treatment needs; thus DoubleU-Net’s superior performance aligns most closely with clinically actionable segmentation.

\subsection{Computational efficiency analysis}
Table~\ref{Table3} compares the computational efficiency of the evaluated models in terms of GFLOPs, parameter count, and inference speed. Here, Mamba-based architectures exhibit lower GFLOPs and smaller parameter counts overall, with VMUNet of 9.25 GFLOPS, VMUNetv2 of 9.90 GFLOPs. However, they have limited boundary precision, especially in RMAMamba-S.  ResUNet++, ColonsegNet and MambaUNet are the smallest model with least number of parameters, with 4.06M, 5.01 M and 15.48M parameters, respectively. 

U-Net achieves the fastest inference at 40.83 FPS. ResUNet++ and DoubleUNet achieve second and third with FPS of 34.22 and 33.09, respectively.  Thus CNN based architecture exhibits strong efficiency, making it well-suited for real-time clinical workflows. Overall, the efficiency results reinforce that CNN-based models provide the most favorable trade-off between accuracy and computational cost. 

\subsection{ROC Analysis}
Figure~\ref{fig:roccurvecomparision} shows the ROC curve for all the architectures. Here,  all models exhibit high discriminative ability $(AUC > 0.87)$, indicating that all the methods are capable of reliably classifying individual pixels as carious or non-carious. Among the CNN-based architectures, \textit{DoubleUNet} obtained the highest AUC score of $0.9467$. In the Transformer family, \textit{PVTFormer} achieved the highest AUC of $0.9269$. Interestingly, Mamba-based architecture performed impressively well with VMUNet obtaining AUC of $0.9782$, VmUNetV2 of $0.9782$ and MambaUNet obtaining the overall highest AUC of $0.9816$.

However, a high AUC does not guarantee accurate segmentation, as segmentation quality depends not only on discriminability but also on the model’s ability to produce anatomically meaningful masks. This distinction is important in medical imaging. It explains why Mamba and Transformer models, despite better AUCs, struggle with mDSC and mIoU. Their prediction often shows oversegmentation and fails to capture lesion counters as compared to CNN-based methods. Here, DoubleU-Net and U-Net are able to translate their discriminative ability into much sharper and clinically consistent mask boundaries, maintaining a strong balance between true positives and false positives in spatial context.

\subsection{Key limitation}
Although this research offers a detailed benchmark of the DC1000 dataset, it has certain limitations. While the panoramic radiographs are diverse, they remain relatively small compared to the natural image benchmark. This might be one of the possible reasons why Transformers and Mamba-based models are performing low. The dataset also contains class imbalance and numerous low-contrast or subtle lesions, which may have limited the performance.

\section{Conclusion}
This work presents a comprehensive analysis of twelve SOTA CNN, Transformer, and Mamba-based architectures for dental caries segmentation on the DC1000 panoramic radiograph dataset. Among all the models, the CNN-based DoubleU-Net achieved the highest overall mDSC of 0.7345, mIoU of 0.5978, and Precision of 0.8145. In contrast, Transformer and Mamba variants, despite strong discriminative ability reflected in high AUC values, often fail to translate this into precise boundary localization. These results highlight that architectural success in medical imaging depends less on model complexity and more on spatial inductive priors, data scale, and the ability to capture fine-grained texture cues. The computational analysis further shows the practicality of CNNs, which showed more favorable trade-offs than their Transformer and Mamba counterparts. Future work includes developing more datasets in the field through multi-institutional data collection, and expanding the problem to more classes, such as impacted tooth, periapical radiolucency, etc, in addition to caries and testing our algorithm on more heterogeneous cohorts. 

\bibliography{aaai2026}

\end{document}